\documentclass[10pt,twocolumn]{article}

\usepackage[letterpaper,left=0.875in,right=0.875in,top=1in,bottom=1in]{geometry}
\usepackage{times}
\usepackage[T1]{fontenc}
\usepackage{microtype}
\usepackage{fancyhdr}
\fancypagestyle{plain}{\fancyhf{}\fancyfoot[C]{\small\thepage}}
\usepackage{titlesec}
\titleformat{\section}{\normalsize\bfseries}{\thesection}{0.6em}{}
\titleformat{\subsection}{\normalsize\bfseries}{\thesubsection}{0.6em}{}
\titleformat{\paragraph}[runin]{\normalsize\bfseries}{}{0pt}{}[.]
\titlespacing*{\section}{0pt}{10pt plus 2pt}{4pt}
\titlespacing*{\subsection}{0pt}{7pt plus 2pt}{3pt}
\titlespacing*{\paragraph}{0pt}{5pt}{0.6em}

\usepackage{amsmath}
\usepackage{booktabs}
\usepackage{array}
\usepackage{graphicx}
\usepackage{xcolor}
\usepackage{enumitem}
\usepackage[font=small,labelfont=bf,skip=4pt]{caption}
\usepackage{natbib}
\usepackage[colorlinks=true,linkcolor=blue!50!black,citecolor=blue!50!black,urlcolor=blue!50!black]{hyperref}
\setcitestyle{authoryear,round,aysep={},yysep={;}}
\setlist{nosep,leftmargin=1.4em}

\begin{document}

\twocolumn[
\vspace*{-2pt}
\hrule height 1.2pt
\vskip 10pt
\begin{center}
{\Large\bfseries MarineSparkLM-4B: A Lightweight Language Model for Marine and Offshore Engineering via Parameter-Efficient Adaptation\par}
\end{center}
\vskip 6pt
\hrule height 1.2pt
\vskip 12pt
\begin{center}
{\large\bfseries Cheng Siong Chin}\par
\vskip 3pt
Faculty of Science, Agriculture and Engineering, Newcastle University Singapore\par
\texttt{cheng.chin@newcastle.ac.uk}\par
\end{center}
\vskip 8pt
\begin{center}
\begin{minipage}{0.88\textwidth}
\begin{center}{\bfseries Abstract}\end{center}
\vskip -4pt
\small
Marine and offshore engineering text is spread across many sources and is small by pretraining standards, so training a domain model from scratch is not practical. We study how far low-rank adaptation (LoRA) can take a 4-billion-parameter ocean-domain model, OceanGPT-basic-4B-Instruct, on a single NVIDIA DGX Spark. Two independent adapters, each training 0.81\% of the parameters, are compared. The first is trained by next-token prediction on a 61.6-million-token corpus of 7,069 documents in ten engineering themes, built with deduplication and a two-tier keyword filter that removes cross-domain lexical collisions. The second is trained on 2,295 question-answer pairs with the loss on answer tokens only, after the first adapter drifted off topic on a retrieval-grounded ballast-water question. On two sets of 20 hand-scored prompts, the instruction adapter outperforms both the base model and the continuation adapter (paired Wilcoxon $p\le 0.0008$ on the first set and $p\le 0.0002$ on a second set built from passages unseen in training, validation and checkpoint selection). Retrieval improves the base model ($p=0.0021$) and adds no detectable gain to the instruction adapter. Accuracy remains modest, at 8 of 40 on the second set. A cheap automatic metric, embedding similarity to the reference answer, correlates moderately with the human scores ($r=0.55$, $p<10^{-18}$, $n=220$, reusing generations already collected), and the instruction adapter adds 0.65 s of load time and negligible memory over the base model on one device. Further checks find no benefit from LoRA rank beyond 8, no catastrophic forgetting on MMLU (Massively Multitask Language Understanding), a same-size generic instruct baseline (Qwen3-4B-Instruct-2507, OceanGPT's own base) that beats the bare base model but trails our instruction adapter, no meaningful bias in blind re-scoring, and that 4-bit quantisation cuts memory by 62\% with comparable quality. An enlarged 120-question set confirms the central result at 6$\times$ scale while exposing a real limit of the automatic metric, caught only by a hand-scored spot-check. We report the full pipeline, a reverted filter change, an instruction run that overfit within 100 steps, and the limits of the evaluation.
\end{minipage}
\end{center}
\vskip 10pt
]

\section{Introduction}

Language models adapted to one technical field help where general-purpose models lack vocabulary and context. SciBERT and BioBERT were pretrained on scientific and biomedical text \citep{beltagy2019scibert,lee2020biobert}, and BloombergGPT on a large financial corpus \citep{wu2023bloomberggpt}. \citet{gururangan2020dont} showed that continued pretraining on in-domain text improves downstream performance even for models that already had large general pretraining.

Marine and offshore engineering has received little of this attention. Its literature covers hull design, wave loads, propulsion, offshore structures, offshore energy, control, uncrewed vehicles, underwater sensing, numerical simulation and shipping operations. Three published models address the marine domain from different angles. OceanGPT targets ocean science tasks \citep{bi2023oceangpt}. MarineGPT is a vision-language model trained on more than five million marine image-text pairs \citep{zheng2023marinegpt}. Llamarine continues the pretraining of Llama 3.1 70B on maritime text for navigation tasks \citep{nguyen2025llamarine}. In our searches of arXiv and the open web we did not find a published text-only model adapted on a corpus organised around marine and offshore engineering themes.

Two obstacles stand in the way. The first is corpus size. A common reading of the compute-optimal scaling results of \citet{hoffmann2022chinchilla} is that a model should see about 20 training tokens per parameter. A 500-million-parameter model would need roughly 10 billion tokens, and the corpus we could assemble held about 70 million, which is 142 times fewer. Adapting an existing model is the workable route. The second obstacle is corpus quality. Keyword-based collection admits documents in which a term has an unrelated everyday meaning: in our first filtering pass a Wikipedia biography was kept as a control-systems document because it contained ``control,'' and a toilet-seat product page was kept as an offshore-structures document because it contained ``riser.'' A third problem appeared only after training. An adapter trained by next-token prediction on the corpus still answered a ballast-water treatment question with general marine pollution, even when the prompt contained retrieved passages on exactly that topic.

We target a lightweight model on purpose. Marine engineering groups, shipyards and offshore operators often work on a single workstation, and a model that can be adapted on one device is easier to refresh as documents arrive. Here, lightweight means a 4-billion-parameter base model (about 8 GB of weights), adapters that train 0.81\% of the parameters (66 MB per adapter at 16-bit precision, computed from the parameter count), and training runs of 4 h 40 min and 10 min on one DGX Spark. Section~\ref{sec:autoinf} reports inference cost directly: the instruction adapter adds 0.65 s of load time and no measurable memory over the base model's own 8.31 GB footprint.

\paragraph{Contributions} (i) A ten-theme corpus of 61.6 million tokens and a two-tier keyword filter designed by tracing individual false positives to the keyword responsible. (ii) Two independently trained LoRA adapters on one frozen base model, one by continuation training and one by answer-masked instruction tuning. (iii) Two hand-scored evaluations, the second on prompts built from passages absent from instruction training, validation and checkpoint selection, with paired significance tests. (iv) Negative and null results: an abandoned pretraining plan, a reverted keyword reclassification, a 114-pair run that overfit within 100 steps, no detectable retrieval gain over the instruction adapter, and no detectable benefit from the hand-written pairs. (v) Public release of the code, adapter, retrieval index and evaluation files (Availability).

\section{Related Work}

\paragraph{Domain language models} SciBERT, BioBERT and BloombergGPT show that in-domain text helps encoders and decoders on field-specific tasks \citep{beltagy2019scibert,lee2020biobert,wu2023bloomberggpt}. \citet{gururangan2020dont} separate domain-adaptive pretraining, which uses a large unlabeled in-domain corpus, from task-adaptive pretraining. Our continuation adapter is a low-rank variant of the first kind.

\paragraph{Marine and maritime models} OceanGPT is a family of models for ocean science tasks \citep{bi2023oceangpt}; our base model belongs to this line. MarineGPT aligns vision and language for marine images using Marine-5M \citep{zheng2023marinegpt}. Llamarine starts from Llama 3.1 70B and uses continued pretraining and fine-tuning on maritime textbooks, publications and Wikipedia text \citep{nguyen2025llamarine}. We exclude MarineGPT from the quantitative comparison because our test sets contain text only, and Llamarine because the only checkpoint we located was a 141 GB third-party upload of unverified provenance, too large for our hardware.

\paragraph{Low-rank adaptation} LoRA freezes pretrained weights and trains low-rank update matrices added to selected layers \citep{hu2022lora}. \citet{biderman2024lora} found that LoRA learns less than full fine-tuning on target domains such as code and mathematics and also forgets less of the base model's behaviour elsewhere, a trade-off that suits a small corpus on top of a base model that already carries relevant knowledge.

\paragraph{Synthetic instruction data and retrieval} Self-Instruct showed that a model can generate its own instruction data when the output is filtered \citep{wang2023selfinstruct}. We make one change: the generator reads a real corpus passage, so each pair is tied to a source text. Retrieval-augmented generation conditions a model on retrieved passages at inference time \citep{lewis2020rag}.

\paragraph{Corpus hygiene} Exact and near-duplicate removal improves language models and lowers memorisation \citep{lee2022dedup}. We combine exact hashing with MinHash resemblance detection \citep{broder1997resemblance}.

\section{Method}

\begin{figure*}[t]
\centering
\includegraphics[width=0.92\textwidth]{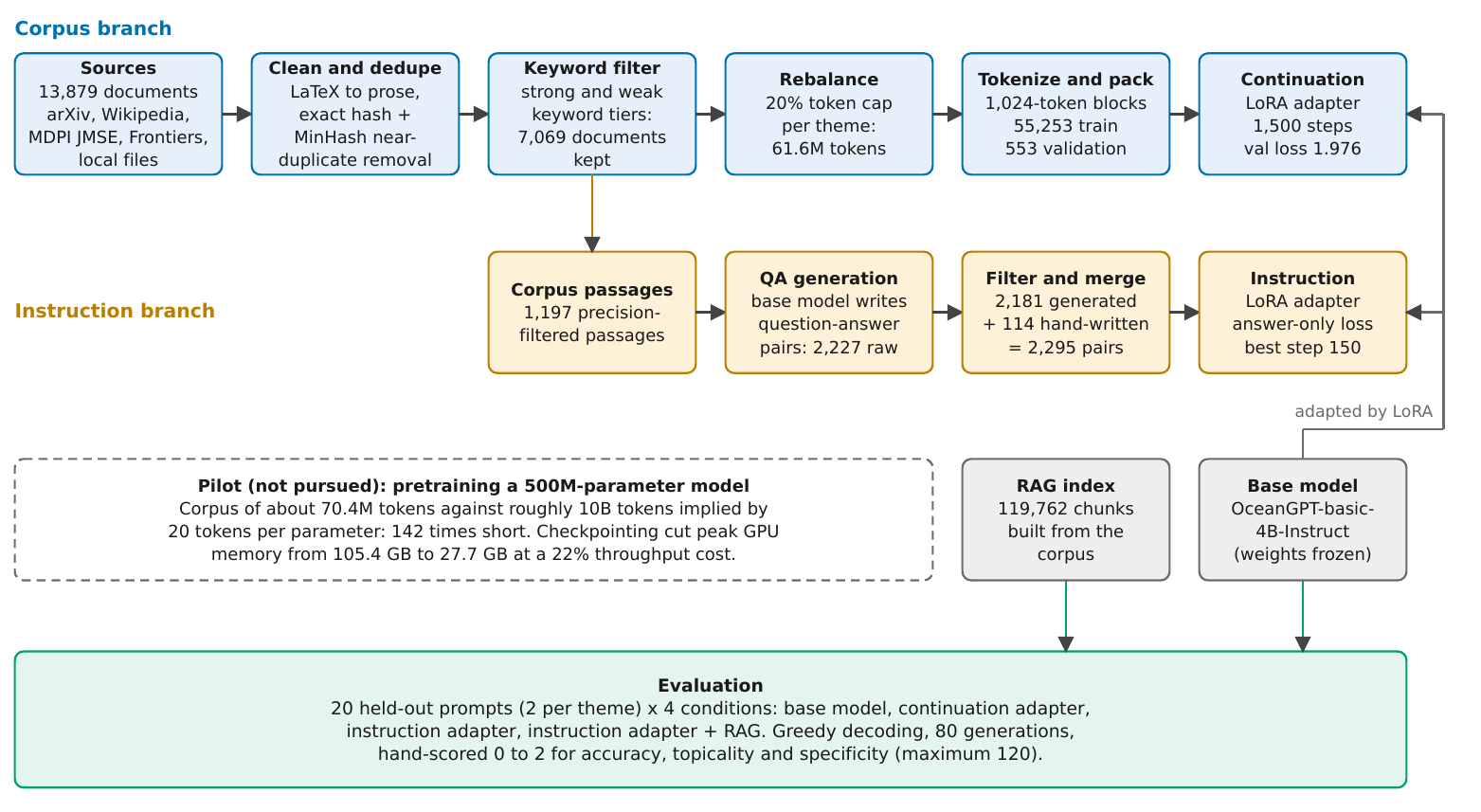}
\caption{Pipeline overview. Two LoRA adapters are trained independently from one frozen base model. The corpus branch yields the continuation adapter and the instruction branch yields the instruction adapter. The retrieval index is built from the same corpus.}
\label{fig:pipeline}
\end{figure*}

\subsection{Base model and the decision not to pretrain}

The base model is OceanGPT-basic-4B-Instruct (Hugging Face identifier \texttt{zjunlp/OceanGPT-basic-4B-Instruct}, MIT licence), a text-only model with the Qwen3 decoder architecture \citep{yang2025qwen3}. It has 36 layers, hidden size 2,560, 32 attention heads with 8 key-value heads, a vocabulary of 151,936, and 4,055,498,240 parameters, and runs in bfloat16.

Before settling on adaptation we benchmarked a 500-million-parameter decoder-only pretraining configuration on one DGX Spark (GB10, unified memory, ARM64). Without gradient checkpointing \citep{chen2016checkpointing} the run used 105.4 GB of GPU memory at steady state and sustained 9,548 tokens per second. With checkpointing it used 27.7 GB at 7,434 tokens per second, a 22\% throughput cost. The benchmark covered iterations 0 to 15 and served only to size a full run. The corpus decided the matter: the shortfall against 10 billion tokens ranged from 1,327 times on the smaller corpora we measured to 142 times on the merged 70.4-million-token corpus. We did not run the full pretraining.

\subsection{Corpus construction}

The corpus covers ten themes: naval architecture, hydrodynamics, propulsion and machinery, offshore structures, offshore energy, control and autonomy, marine vehicles, sensors and communications, simulation and modelling, and maritime operations. Documents came from arXiv, Wikipedia, the MDPI \emph{Journal of Marine Science and Engineering}, Frontiers journals, and local technical documents in PDF, DOCX, PPTX and HTML formats. Cleaning converts LaTeX source to prose and repairs text encoding. Deduplication removes exact copies by hash and near-duplicates by MinHash. Keyword filtering keeps a document for a theme only if it passes the tier rule below. Rebalancing caps each theme at about 20\% of the tokens. For fine-tuning, the base tokenizer converts documents to a single stream that is cut into 1,024-token blocks, and validation documents are held out by document. Successive cleaning passes reduced the corpus from 70.4 million tokens to the final 61.6 million.

\paragraph{Keyword tiers} The first filter kept a document if any theme keyword appeared anywhere in it. We traced each off-topic document to the keyword that admitted it and split every theme's keyword list into two tiers. The strong tier holds specific, low-collision terms and multi-word phrases. The weak tier holds common English words with demonstrated or plausible collisions, such as ``control'' and ``riser.'' A document is kept if it matches one strong keyword or several independent weak keywords. For two themes, moving further keywords to the weak tier removed more genuine engineering content (a bow-thruster field study and a North Sea oil and gas field list) than contamination, so those reclassifications were reverted. One off-topic document, a fitness article matching ``thruster,'' remains in the propulsion theme for the same reason.

Of 13,879 raw documents, 7,069 (50.9\%) were kept, giving 61,573,332 tokens under the project tokenizer; theme shares range from 6.35\% to 14.06\%, so no theme reached the 20\% cap (Appendix~\ref{app:corpus}). Under the base tokenizer the training and validation splits hold 56,579,072 and 566,272 tokens (55,253 and 553 blocks).

\subsection{Adapters}

Both adapters share one LoRA configuration: rank 16, scaling factor 32 and dropout 0.05 on the \texttt{q\_proj}, \texttt{k\_proj}, \texttt{v\_proj}, \texttt{o\_proj}, \texttt{gate\_proj}, \texttt{up\_proj} and \texttt{down\_proj} modules. This trains 33,030,144 parameters, 0.81\% of the base model, a deliberately small fraction given about 56.6 million packed training tokens. Each adapter is trained from the base model directly; the two are never stacked. Training used AdamW \citep{loshchilov2019adamw} through the Hugging Face \texttt{Trainer} \citep{wolf2020transformers} with weight decay 0.1 and gradient-norm clipping at 1.0; Appendix~\ref{app:hparams} lists the settings.

\paragraph{Continuation adapter} It is trained by next-token prediction on the packed corpus for 1,500 steps (effective batch of 8 sequences, 8,192 tokens per step).

\paragraph{Instruction adapter} The instruction set has 2,295 question-answer pairs over the ten themes. A hand-authored subset of 114 pairs was written by reading a corpus passage and writing a question it answers. The other 2,181 pairs were generated by prompting the unadapted base model, to avoid feeding one adapter's biases into the other, to write JSON question-answer pairs from 1,197 further precision-filtered passages. Generation gave 2,227 raw pairs, after 72 of the 1,197 passages failed to yield parseable JSON. Of these 2,227, 14 were dropped as journal-metadata questions and 32 were dropped as cross-domain collisions such as a spacecraft-thruster paper and an orbital-mechanics article on tidal acceleration, leaving 2,181. The loss is applied to answer tokens only, and each example is padded or truncated to 512 tokens instead of packed, since packing would blur the masking boundary between unrelated pairs. A validation split of 229 pairs was held out. Training on the 114 hand-written pairs alone overfit at once: training loss fell from 3.70 to 0.02 by step 100 while validation loss rose from a best of 2.45 at step 25 to 3.89. The full set stopped early at step 300 of a 600-step budget, with the best checkpoint at step 150 (validation loss 1.853).

\subsection{Retrieval}

The default retriever is TF-IDF (Term Frequency--Inverse Document Frequency) over an index of 119,762 chunks built from the corpus, and its passages are used to condition generation. A dense alternative embeds the same chunks with BAAI/bge-small-en-v1.5 and returns no passage whose similarity falls below a calibrated floor.

\section{Experiments}

\subsection{Setup}

\paragraph{Conditions} We compare the base model, the continuation adapter, the instruction adapter and the instruction adapter with retrieval. The second evaluation adds the base model with TF-IDF retrieval, the instruction adapter with dense retrieval, and a synthetic-only adapter trained on the 2,181 generated pairs without the 114 hand-written ones. All conditions use greedy decoding.

\paragraph{Prompt sets} Evaluation A has 20 prompts, two per theme, drawn from the 229-pair validation split with the seed used at training time. Because this split also selected the instruction checkpoint through validation loss, we built Evaluation B from corpus passages whose document identifiers appear nowhere in the instruction set, neither in its training nor its validation portion, giving 20 questions never seen in training, validation or checkpoint selection. In both evaluations the retrieval function excludes the source document of the prompt being answered, since the index contains that document; we verified this exclusion is applied in Evaluation B for both the TF-IDF and dense retrieval conditions.

\paragraph{Scoring and statistics} Each generation is scored by hand against its reference answer on accuracy (agreement with the facts in the reference), topicality (staying on the correct marine engineering subtopic) and specificity (concrete domain detail instead of generic filler), each from 0 to 2. A condition's maximum is 40 per dimension and 120 in total, and a per-item total ranges from 0 to 6. We report paired Wilcoxon signed-rank tests and bootstrap 95\% confidence intervals (10,000 resamples) on the mean paired difference in per-item totals. At $n=20$ these indicate the direction of an effect and do not support strong significance claims. Reference answers were written by the base model from the source passages.

\subsection{Training results}

\begin{figure}[t]
\centering
\includegraphics[width=0.92\columnwidth]{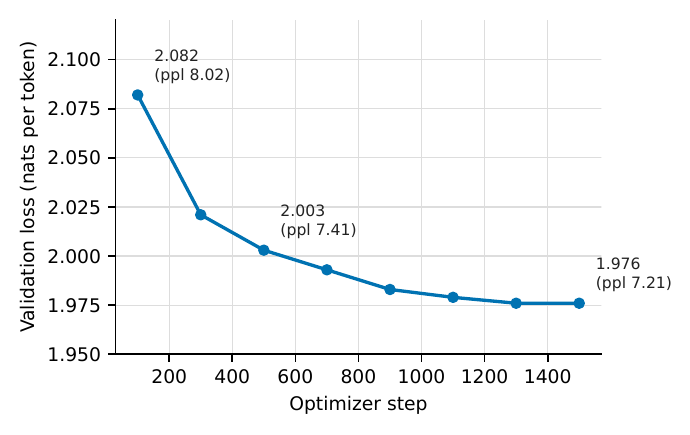}
\caption{Validation loss of the continuation adapter at the eight reported evaluation steps.}
\label{fig:loss}
\end{figure}

Validation loss of the continuation adapter fell from 2.082 (perplexity 8.02) at step 100 to 1.976 (7.21) at step 1,500 without a plateau or reversal (Figure~\ref{fig:loss}); early stopping never triggered and the run-mean training loss was 1.987. Loss before training was not recorded on this split, so the contribution of the first 100 steps is unknown. Completions for ten fixed prompts, one per theme, were coherent, on topic and free of repetition collapse. Loss values of the two adapters are not comparable, because they are computed over different token sets.

\subsection{Evaluation A}

\begin{figure}[t]
\centering
\includegraphics[width=0.98\columnwidth]{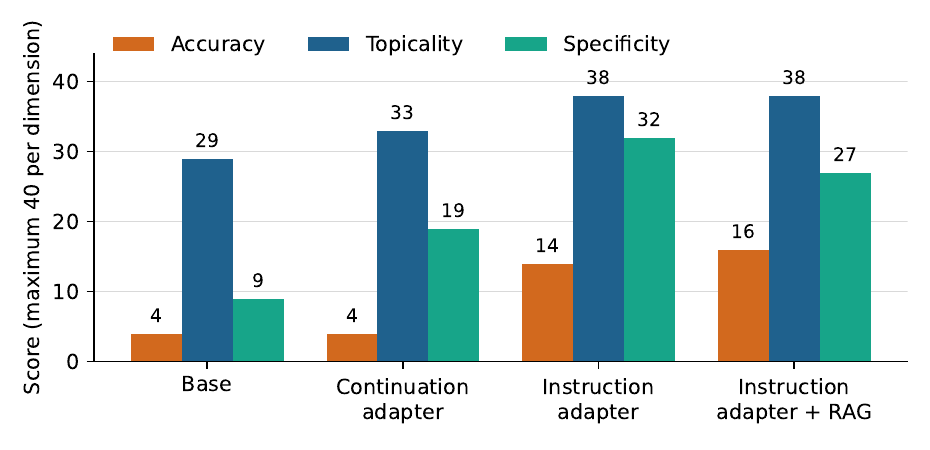}
\caption{Evaluation A scores by dimension for the four conditions.}
\label{fig:scores}
\end{figure}

Table~\ref{tab:scoresA} and Figure~\ref{fig:scores} give the scores. The instruction adapter reaches 84 of 120 against 42 for the base model and 56 for the continuation adapter, driven by specificity (9, 19, 32) and accuracy (4, 4, 14). Both gains hold under paired tests (Table~\ref{tab:stats}). Adding retrieval leaves the total essentially unchanged (84 without, 81 with; paired difference $-0.15$, 95\% CI $[-0.50,+0.20]$, $p=0.41$), with accuracy up from 14 to 16 and specificity down from 32 to 27. Topicality is high in every condition and rises from 29 to 38. On 2 of the 20 items the retrieval output matched the output without retrieval, which indicates that retrieval returned nothing useful for those vaguely worded queries.

\begin{table}[t]
\centering
\caption{Evaluation A: hand-scored comparison on 20 prompts (maximum 40 per dimension, 120 in total).}
\label{tab:scoresA}
\small
\begin{tabular}{@{}lrrrr@{}}
\toprule
Condition & Acc. & Top. & Spec. & Total \\
\midrule
Base model & 4 & 29 & 9 & 42 \\
Continuation adapter & 4 & 33 & 19 & 56 \\
Instruction adapter & 14 & 38 & 32 & 84 \\
Instruction adapter + RAG & 16 & 38 & 27 & 81 \\
\bottomrule
\end{tabular}
\end{table}

\subsection{Evaluation B}

\begin{table}[t]
\centering
\caption{Evaluation B: hand-scored comparison on 20 prompts never used in training, validation or checkpoint selection (maximum 40 per dimension, 120 in total). TF-IDF and dense name the retriever.}
\label{tab:scoresB}
\small
\begin{tabular}{@{}lrrrr@{}}
\toprule
Condition & Acc. & Top. & Spec. & Total \\
\midrule
Base model & 3 & 24 & 14 & 41 \\
Continuation adapter & 1 & 28 & 22 & 51 \\
Base + RAG (TF-IDF) & 5 & 32 & 18 & 55 \\
Instruction + RAG (TF-IDF) & 9 & 34 & 30 & 73 \\
Instruction adapter & 8 & 40 & 31 & 79 \\
Instruction + RAG (dense) & 8 & 40 & 30 & 78 \\
Instruction, synthetic-only & 9 & 40 & 33 & 82 \\
\bottomrule
\end{tabular}
\end{table}

The central result holds on the clean set (Table~\ref{tab:scoresB}, Table~\ref{tab:stats}). Retrieval significantly helps the base model on its own (mean difference $+0.70$, $p=0.0021$) but shows no significant effect on the instruction adapter with either retriever ($p=0.08$ for TF-IDF, trending negative, and $p=0.32$ for dense, roughly neutral). One reading is that instruction tuning already supplies much of what retrieval added on this prompt mix, topical anchoring and commitment to a concrete answer, leaving little for retrieval to add. The two retrievers do not differ significantly overall ($p=0.17$), but one item illustrates the quality gap. For an offshore-structures prompt about a coupled mooring and viscous-flow solver, TF-IDF retrieved an unrelated passage on four-parameter semitoric systems, a pure-mathematics topic and most likely a keyword collision, and the model produced a fluent but falsely grounded answer. Dense retrieval returned nothing above its floor, so the model fell back on its own on-topic though still incorrect answer.

The synthetic-only adapter scores 82 against 79 for the full mix, a difference that is not significant (mean $+0.15$, 95\% CI $[-0.15,+0.50]$, $p=0.33$). The hand-written pairs were needed to expose the overfitting failure that motivated the scale-up, but their presence in the final mix is not distinguishable from their absence on this sample.

\begin{table*}[t]
\centering
\caption{Paired comparisons of per-item totals (scale 0 to 6): mean difference, bootstrap 95\% CI and Wilcoxon $p$-value. The first group is Evaluation A and the second is Evaluation B.}
\label{tab:stats}
\small
\begin{tabular}{@{}llrrr@{}}
\toprule
Eval. & Comparison (second minus first) & Mean diff. & 95\% CI & $p$ \\
\midrule
A & Base $\to$ instruction adapter & $+2.10$ & $[+1.40,+2.75]$ & 0.0005 \\
A & Continuation $\to$ instruction adapter & $+1.40$ & $[+0.85,+1.95]$ & 0.0008 \\
A & Instruction adapter $\to$ with RAG (TF-IDF) & $-0.15$ & $[-0.50,+0.20]$ & 0.4054 \\
\midrule
B & Base $\to$ instruction adapter & $+1.90$ & $[+1.45,+2.25]$ & 0.0001 \\
B & Continuation $\to$ instruction adapter & $+1.40$ & $[+1.00,+1.85]$ & 0.0002 \\
B & Base $\to$ base + RAG (TF-IDF) & $+0.70$ & $[+0.40,+1.00]$ & 0.0021 \\
B & Instruction adapter $\to$ with RAG (TF-IDF) & \multicolumn{2}{c}{(see text)} & 0.08 \\
B & Instruction adapter $\to$ with RAG (dense) & \multicolumn{2}{c}{(see text)} & 0.32 \\
B & TF-IDF $\to$ dense retrieval, instruction adapter & \multicolumn{2}{c}{(see text)} & 0.17 \\
B & Full mix $\to$ synthetic-only adapter & $+0.15$ & $[-0.15,+0.50]$ & 0.3340 \\
\bottomrule
\end{tabular}
\end{table*}

\subsection{Qualitative observations}

The motivating failure was a ballast-water treatment prompt: even with retrieved passages on the exact topic, the continuation adapter drifted toward general marine pollution. After instruction tuning the same prompt stayed on topic and cited International Maritime Organization discharge standards, and a dynamic-positioning prompt and a new double-hull prompt also stayed on topic. Through the deployed web interface (Appendix~\ref{app:interface}), a reworded ballast-water prompt and an ROV (remotely operated vehicle) thruster-allocation prompt stayed on topic in direct mode, and a mooring-tension prompt in retrieval mode retrieved a passage from the offshore-structures theme (retrieval score 0.365) and produced a grounded answer.

On the vague prompt ``What is the primary goal of the research described in the document?'' (sensors and communications), the base model answered about anti-cancer marine natural compounds and the continuation adapter about the thermal performance of building insulation, while both instruction conditions stayed within marine engineering (autonomous underwater vehicle control) without giving the correct answer. One item defeated all four Evaluation A conditions: asked which metric was used to optimise a neural network (reference answer: balanced accuracy, chosen for unbalanced classes), every condition named mean squared error or a generic loss. That fact appears in one paper's methodology and in no training split. A query about a named individual outside the corpus drew a fluent biography with a fabricated affiliation, a reminder that the adapters make the model commit to concrete answers whether or not they are correct.

\subsection{Automatic metric and inference cost}\label{sec:autoinf}

To check whether a cheap automatic metric could reduce the hand-scoring burden at larger scale, we computed embedding cosine similarity (BAAI/bge-small-en-v1.5) between each generation and its reference answer and correlated it with the human per-item total, reusing the generations and scores already collected rather than generating anything new. The correlation is moderate and significant: Pearson $r=0.34$ ($p=0.002$, $n=80$) on Evaluation A, $r=0.67$ ($p<10^{-4}$, $n=140$) on Evaluation B, and $r=0.55$ ($p<10^{-18}$, $n=220$) combined. This is not strong enough to replace hand scoring outright, but strong enough to triage a larger prompt set, for example by sending only the largest automatic-human disagreements to a human rater instead of scoring every item by hand.

We also measured single-prompt inference cost for the instruction adapter on one DGX Spark GB10 node (Table~\ref{tab:inference}). The adapter adds negligible load time and no measurable memory over the base model, consistent with the lightweight-adaptation framing. We did not measure batched throughput or a consumer GPU; Section~\ref{sec:robustness} reports a quantised variant.

\begin{table}[t]
\centering
\caption{Inference cost, instruction adapter, one DGX Spark GB10, single prompt, greedy decoding.}
\label{tab:inference}
\small
\begin{tabular}{@{}lr@{}}
\toprule
Measurement & Value \\
\midrule
Base model load time & 43.6 s \\
Adapter load time & 0.65 s \\
Peak GPU memory, load & 8.31 GB \\
Peak GPU memory, generation & 8.22 GB \\
Throughput & 17.6 tokens/s \\
\bottomrule
\end{tabular}
\end{table}

\subsection{Robustness and cost checks}\label{sec:robustness}

Five further checks probe adapter capacity, general-capability retention, an external baseline, rater reliability, and deployment cost.

\textbf{LoRA rank sweep.} Training the instruction adapter at rank 8 and rank 64, with alpha fixed at 32 and all other hyperparameters unchanged, gives best validation loss 1.853 (rank 8, step 150), 1.853 (rank 16, production, step 150), and 1.850 (rank 64, step 250). The three ranks converge to essentially the same loss; we find no benefit from capacity beyond rank 8 at this dataset size, consistent with data rather than capacity being the bottleneck.

\textbf{Catastrophic forgetting.} We scored the base model and instruction adapter on the same random 200-question MMLU subset \citep{hendrycks2021mmlu}, comparing next-token logits for the four answer-letter tokens directly rather than generating free text and regex-parsing it, which gave an implausible result on an initial smoke test. The base model scored 117/200 (0.585) and the instruction adapter 126/200 (0.630), a delta of $+9$ ($+0.045$). We detect no forgetting.

\textbf{External baseline.} OceanGPT-basic-4B-Instruct is itself built from Qwen3-4B-Instruct-2507 \citep{yang2025qwen3}. Running the plain Qwen3-4B-Instruct-2507 through the same harness against Evaluation B, with and without TF-IDF RAG, scores 56/120 and 55/120 respectively, against 41 for bare OceanGPT, 51 for the continuation adapter, and 79 for our instruction adapter (Table~\ref{tab:scoresB}). Plain Qwen3 beats bare OceanGPT and matches the continuation adapter and base+RAG, but remains far below our instruction adapter; it was also qualitatively more calibrated, often declining to answer when a retrieved passage did not address the question rather than hallucinating. This complicates the assumption that OceanGPT's domain pretraining was itself a clear improvement over a same-size generic base, while confirming that our instruction tuning contributes large value regardless of the base model.

\textbf{Rater reliability.} Blind re-scoring of a stratified sample of 40 generations (20 from each evaluation), with condition and original score hidden, agrees exactly with the original score on 34/40 items (0.850), mean absolute difference 0.150, Pearson $r=0.969$ ($p=1.0\times10^{-24}$), net bias $-4$ points across 40 items. We find no meaningful rater bias.

\textbf{Quantisation.} Loading the instruction adapter in 4-bit NF4 (NormalFloat4) via bitsandbytes \citep{dettmers2023qlora} instead of bf16 changes load time from 71.0\,s to 74.6\,s, peak memory from 8.31/8.22\,GB (load/generation) to 3.12/3.03\,GB (a 62\% reduction), and throughput from 7.9 to 15.1 tokens/s, with comparable output quality on the sample prompt. The 4-bit model fits comfortably on an 8\,GB consumer GPU.

\subsection{Enlarged held-out set}\label{sec:round4icml}

Evaluation B was 20 questions. We enlarged it to 120 (12/theme, 6$\times$), re-running the 4 core conditions and scoring with the automatic metric from Section~\ref{sec:autoinf} rather than hand-scoring 480 generations; the 3 ablation conditions added in Evaluation B already got null results at $n=20$ and were not re-run. The set was built the same leakage-free way, additionally excluding every passage already in Evaluation B; 160 candidate passages yielded 287 raw pairs, an automated filter and one-question-per-document rule gave 120, and a manual pass caught 10 keyword-collision false positives and one OCR-garbled item, each replaced.

\begin{table}[t]
\centering
\caption{Enlarged set (n=120): mean automatic-metric score, bootstrap 95\% CI.}
\label{tab:round4icml}
\small
\begin{tabular}{@{}lrr@{}}
\toprule
Condition & Mean & 95\% CI \\
\midrule
Base & 0.7324 & [0.7119, 0.7529] \\
Continuation & 0.7000 & [0.6815, 0.7191] \\
Instruction & 0.7596 & [0.7432, 0.7755] \\
Instruction + RAG (TF-IDF) & 0.7461 & [0.7272, 0.7647] \\
\bottomrule
\end{tabular}
\end{table}

The central result replicates at 6$\times$ scale (base $\to$ instruction: $p=0.0118$; continuation $\to$ instruction: $p=4.5\times10^{-8}$), and RAG still shows no significant benefit on the instruction adapter ($p=0.24$). But the base-vs-continuation row is wrong: the metric says base beats continuation ($p=4.4\times10^{-5}$), the opposite of Evaluations A/B's hand scoring. A 24-item hand-scored spot-check confirms continuation beats base (3.00/6 vs.\ 2.17/6, the correct direction) despite good item-level correlation on the same spot-check ($r=0.619$, $\rho=0.680$). The likely cause: reference answers were written by the base model, so base's own phrasing resembles them more closely regardless of correctness, a bias that only surfaces against the model that wrote the references.

\subsection{Author-reviewed references on a 30-item subset}\label{sec:round5icml}

Every reference answer so far was written by the base model. To check whether this inflated the instruction adapter's advantage, the author independently drafted a grounded answer for a stratified 30-item subset of the enlarged set (3/theme), reading only the source passage, not the model's existing reference. This doubled as a QC pass: it caught 4 of the 30 items as off-topic despite passing the enlarged set's own manual read-through (a road-vehicle control passage, a bacterial-swimming biophysics passage, a spacecraft-propulsion passage, and a supersonic cavity-flow passage, each misfiled under an unrelated marine theme), plus one borderline item. The author chose to keep all 30 as drawn rather than replace them.

\begin{table}[t]
\centering
\caption{30-item subset: mean automatic-metric score, model-written vs.\ author-reviewed reference.}
\label{tab:round5icml}
\small
\begin{tabular}{@{}lrr@{}}
\toprule
Condition & Model ref. & Author ref. \\
\midrule
Base & 0.7129 & 0.6600 \\
Continuation & 0.6755 & 0.6329 \\
Instruction & 0.7642 & 0.7058 \\
Instruction + RAG (TF-IDF) & 0.7346 & 0.6783 \\
\bottomrule
\end{tabular}
\end{table}

Scores drop by 0.04--0.06 uniformly against the author's shorter, more direct answers, but the central comparisons hold: base $\to$ instruction, $+0.0458$ ($p=0.0405$); continuation $\to$ instruction, $+0.0729$ ($p=9.9\times10^{-5}$); instruction $\to$ RAG, $-0.0275$ ($p=0.33$, no significant benefit). This is a partial check (30 of 120 items, one reviewer, drafts accepted as written), but it is the first round not self-referencing the model under evaluation, and the mechanism that caught the 4-item contamination above.

\section{Discussion}

Instruction tuning produced a clear gain in topicality and specificity on both prompt sets, and the gain survives on prompts that never touched training or model selection. Accuracy is still low (14 of 40 on Evaluation A and 8 of 40 on Evaluation B for the instruction adapter), so the adapters bias the model toward the domain and toward concrete answers without showing that it has reliably learned the facts of the field. The base model wrote 95\% of the instruction pairs, so the instruction adapter is trained largely on that model's own passage-conditioned outputs. We read the gain as better task format and topical anchoring rather than new knowledge, although the study does not isolate the two.

The retrieval results are mixed by design of the evidence. Retrieval helps the plain base model, adds no detectable benefit to the instruction adapter, and its quality depends on the retriever, as the semitoric-systems item shows. Because the index covers the same corpus as the test passages, these evaluations are a restrictive setting for retrieval, and they cannot show how retrieval would perform on arbitrary user queries.

\section{Limitations}

Each evaluation has 20 prompts, scores come from one scoring pass without a second rater, and no inter-rater agreement was measured, so confidence intervals are wide and tests have limited power. Reference answers were written by the same base model that serves as a baseline; Section~\ref{sec:round5icml} checks this on a 30-item subset with author-written references and finds the central comparisons hold, but covers a quarter of the enlarged set with a single reviewer. Some prompts cannot be answered without the source document. The comparisons do not establish whether the passages behind the held-out questions also appear in the continuation corpus, which could favour the continuation adapter. The corpus holds about 56.6 million packed training tokens for a 4-billion-parameter model; LoRA lowers the risk of overfitting but does not remove it, as the 114-pair run showed. Sources carry different licences, and the corpus is not released as a standalone dataset, although the released retrieval index contains chunk text drawn from it. Residual off-topic documents remain, and a low single-digit percentage of generated pairs may carry topical noise from the same keyword collisions; Section~\ref{sec:round5icml}'s closer read of a 30-item subset found 4 (13\%), suggesting the true rate elsewhere in the corpus may run higher. The validation set of 553 blocks is modest. The adapters were not combined, and MarineGPT and Llamarine were not compared. We measured single-prompt load time, memory and throughput on one device (Section~\ref{sec:autoinf}) and a 4-bit quantised variant (Section~\ref{sec:robustness}), but not batched throughput or a consumer GPU directly. The automatic-metric check (Section~\ref{sec:autoinf}) reuses the same small, single-rater evaluation sets and so inherits their limitations; it is not an independent validation. The robustness checks (Section~\ref{sec:robustness}) use a single MMLU subset, a single external baseline, and a 40-item blind re-score, each a bounded probe rather than an exhaustive study. The enlarged set (Section~\ref{sec:round4icml}) is still drawn from the same corpus and theme scheme, not an independent benchmark, and relies on the automatic metric for 3 of 4 comparisons with hand-scoring limited to a 24-item spot-check; the one comparison where that broke (base vs.\ continuation) was caught only because a spot-check was run at all.

\section*{Impact Statement}

This work adapts a language model to marine and offshore engineering. Its outputs are fluent and often wrong in specific detail, as the accuracy scores and the fabricated-biography example show. The model should be used as a starting point for literature orientation and drafting, and any output informing an engineering, safety or regulatory decision needs independent verification against primary sources. Training and adaptation used one workstation-class device, which limits the compute and energy cost of reproducing or refreshing the model.

\section*{Availability}

{\raggedright The code, the retrieval index and the evaluation files are released at \url{https://github.com/mcschin1/MarineSparkLM-4B} under the MIT licence. The instruction adapter is also hosted at \url{https://huggingface.co/mcschin75/MarineSparkLM-4Bv1}. The base model is public under the MIT licence.\par}

\bibliographystyle{plainnat}

\begin{thebibliography}{20}\small
\providecommand{\natexlab}[1]{#1}

\bibitem[Beltagy et al.(2019)]{beltagy2019scibert} Beltagy, I., Lo, K., and Cohan, A. SciBERT: A pretrained language model for scientific text. \emph{arXiv preprint arXiv:1903.10676}, 2019.

\bibitem[Biderman et al.(2024)]{biderman2024lora} Biderman, D. et al. LoRA learns less and forgets less. \emph{Transactions on Machine Learning Research}, 2024. arXiv:2405.09673.

\bibitem[Bi et al.(2023)]{bi2023oceangpt} Bi, Z., Zhang, N., Xue, Y., Ou, Y., Ji, D., Zheng, G., and Chen, H. OceanGPT: A large language model for ocean science tasks. \emph{arXiv preprint arXiv:2310.02031}, 2023.

\bibitem[Broder(1997)]{broder1997resemblance} Broder, A. Z. On the resemblance and containment of documents. In \emph{Proceedings of Compression and Complexity of Sequences}, pp. 21--29, 1997.

\bibitem[Chen et al.(2016)]{chen2016checkpointing} Chen, T., Xu, B., Zhang, C., and Guestrin, C. Training deep nets with sublinear memory cost. \emph{arXiv preprint arXiv:1604.06174}, 2016.

\bibitem[Gururangan et al.(2020)]{gururangan2020dont} Gururangan, S. et al. Don't stop pretraining: Adapt language models to domains and tasks. In \emph{Proceedings of ACL}, pp. 8342--8360, 2020.

\bibitem[Hoffmann et al.(2022)]{hoffmann2022chinchilla} Hoffmann, J. et al. Training compute-optimal large language models. \emph{arXiv preprint arXiv:2203.15556}, 2022.

\bibitem[Hu et al.(2022)]{hu2022lora} Hu, E. J. et al. LoRA: Low-rank adaptation of large language models. In \emph{International Conference on Learning Representations}, 2022. arXiv:2106.09685.

\bibitem[Lee et al.(2020)]{lee2020biobert} Lee, J., Yoon, W., Kim, S., Kim, D., Kim, S., So, C. H., and Kang, J. BioBERT: A pre-trained biomedical language representation model for biomedical text mining. \emph{Bioinformatics}, 36(4):1234--1240, 2020.

\bibitem[Lee et al.(2022)]{lee2022dedup} Lee, K. et al. Deduplicating training data makes language models better. In \emph{Proceedings of ACL}, 2022. arXiv:2107.06499.

\bibitem[Lewis et al.(2020)]{lewis2020rag} Lewis, P. et al. Retrieval-augmented generation for knowledge-intensive NLP tasks. In \emph{Advances in Neural Information Processing Systems}, volume 33, 2020. arXiv:2005.11401.

\bibitem[Loshchilov \& Hutter(2019)]{loshchilov2019adamw} Loshchilov, I. and Hutter, F. Decoupled weight decay regularization. In \emph{International Conference on Learning Representations}, 2019. arXiv:1711.05101.

\bibitem[Nguyen et al.(2025)]{nguyen2025llamarine} Nguyen, W., Phan, H. A., Kimura, K., Maeno, H., Tanaka, M., Le, Q., Poucher, W., and Nguyen, C. Llamarine: Open-source maritime industry-specific large language model. \emph{arXiv preprint arXiv:2503.00203}, 2025.

\bibitem[Wang et al.(2023)]{wang2023selfinstruct} Wang, Y. et al. Self-Instruct: Aligning language models with self-generated instructions. In \emph{Proceedings of ACL}, 2023. arXiv:2212.10560.

\bibitem[Wolf et al.(2020)]{wolf2020transformers} Wolf, T. et al. Transformers: State-of-the-art natural language processing. In \emph{Proceedings of EMNLP: System Demonstrations}, pp. 38--45, 2020.

\bibitem[Wu et al.(2023)]{wu2023bloomberggpt} Wu, S. et al. BloombergGPT: A large language model for finance. \emph{arXiv preprint arXiv:2303.17564}, 2023.

\bibitem[Yang et al.(2025)]{yang2025qwen3} Yang, A. et al. Qwen3 technical report. \emph{arXiv preprint arXiv:2505.09388}, 2025.

\bibitem[Hendrycks et al.(2021)]{hendrycks2021mmlu} Hendrycks, D. et al. Measuring massive multitask language understanding. In \emph{International Conference on Learning Representations}, 2021. arXiv:2009.03300.

\bibitem[Dettmers et al.(2023)]{dettmers2023qlora} Dettmers, T., Pagnoni, A., Holtzman, A., and Zettlemoyer, L. QLoRA: Efficient finetuning of quantized LLMs. In \emph{Advances in Neural Information Processing Systems}, volume 36, 2023. arXiv:2305.14314.

\bibitem[Zheng et al.(2023)]{zheng2023marinegpt} Zheng, Z., Zhang, J., Vu, T.-A., Diao, S., Tim, Y. H. W., and Yeung, S.-K. MarineGPT: Unlocking secrets of ocean to the public. \emph{arXiv preprint arXiv:2310.13596}, 2023.

\end{thebibliography}

\appendix
\onecolumn
\section{Corpus composition}\label{app:corpus}

\begin{table}[h]
\centering
\caption{Final corpus composition by theme. Token counts use the project tokenizer.}
\small
\begin{tabular}{@{}lrrrr@{}}
\toprule
Theme & Tokens & Share (\%) & Raw docs & Kept docs \\
\midrule
Naval architecture & 6,126,958 & 9.95 & 1,990 & 802 \\
Hydrodynamics & 5,981,479 & 9.71 & 1,349 & 651 \\
Propulsion and machinery & 3,908,509 & 6.35 & 802 & 464 \\
Offshore structures & 5,197,355 & 8.44 & 1,712 & 691 \\
Offshore energy & 6,349,375 & 10.31 & 1,012 & 704 \\
Control and autonomy & 8,658,461 & 14.06 & 2,146 & 908 \\
Marine vehicles & 6,081,808 & 9.88 & 1,075 & 788 \\
Sensors and communications & 8,592,191 & 13.95 & 1,054 & 558 \\
Simulation and modelling & 6,191,138 & 10.05 & 1,131 & 771 \\
Maritime operations & 4,486,058 & 7.29 & 1,608 & 732 \\
\midrule
Total & 61,573,332 & 100.00 & 13,879 & 7,069 \\
\bottomrule
\end{tabular}
\end{table}

\section{Training settings}\label{app:hparams}

\begin{table}[h]
\centering
\caption{Training settings for the two adapters. Both ran on one NVIDIA DGX Spark (GB10).}
\small
\begin{tabular}{@{}lll@{}}
\toprule
Setting & Continuation adapter & Instruction adapter \\
\midrule
Training data & 55,253 blocks of 1,024 tokens & 2,295 question-answer pairs \\
Loss & All tokens & Answer tokens only \\
Sequence handling & Packed & Padded or truncated to 512 \\
Per-device batch $\times$ accumulation & $2 \times 4$ & $4 \times 4$ \\
Learning rate, schedule & $10^{-4}$, cosine & $10^{-4}$, cosine \\
Warmup steps & 50 & 30 \\
Step budget & 1,500 & 600 \\
Evaluation interval & 100 steps & 50 steps \\
Early-stopping patience & 3 evaluations & 3 evaluations \\
Outcome & Ran to 1,500 steps & Stopped at step 300 (best at 150) \\
Wall-clock time & 4 h 40 min 11 s & 10 min 6 s \\
\bottomrule
\end{tabular}
\end{table}

\section{Procedure summary}\label{app:procedure}

\begin{enumerate}[nosep]
\item Pretraining a 500-million-parameter model was ruled out because the corpus was about 142 times too small, so a published model was fine-tuned.
\item OceanGPT-basic-4B-Instruct was chosen as the text-only, ocean-domain base; MarineGPT was rejected as a base because it is a vision-language model.
\item Ten themes were collected, cleaned, deduplicated, keyword-filtered with strong and weak tiers and rebalanced, leaving 7,069 documents and 61.6 million tokens.
\item LoRA (rank 16, alpha 32, dropout 0.05) on all attention and MLP projections trained 33 million parameters.
\item The continuation adapter ran for 1,500 steps in about 4 h 40 min, with validation loss falling from 2.08 to 1.98.
\item A grounded ballast-water prompt still drifted to general marine pollution, so continuation training alone was insufficient.
\item 114 hand-written pairs overfit within 100 steps; the base model generated further pairs from 1,197 filtered passages, giving 2,295 pairs.
\item The instruction adapter (answer-only loss, early stop at step 300, best step 150, validation loss 1.853) took about 10 minutes.
\item The ballast-water case was re-tested directly and through the deployed application, and two sets of 20 held-out prompts were scored.
\end{enumerate}

\section{Demonstration interface}\label{app:interface}

\begin{figure}[h]
\centering
\includegraphics[width=0.7\textwidth]{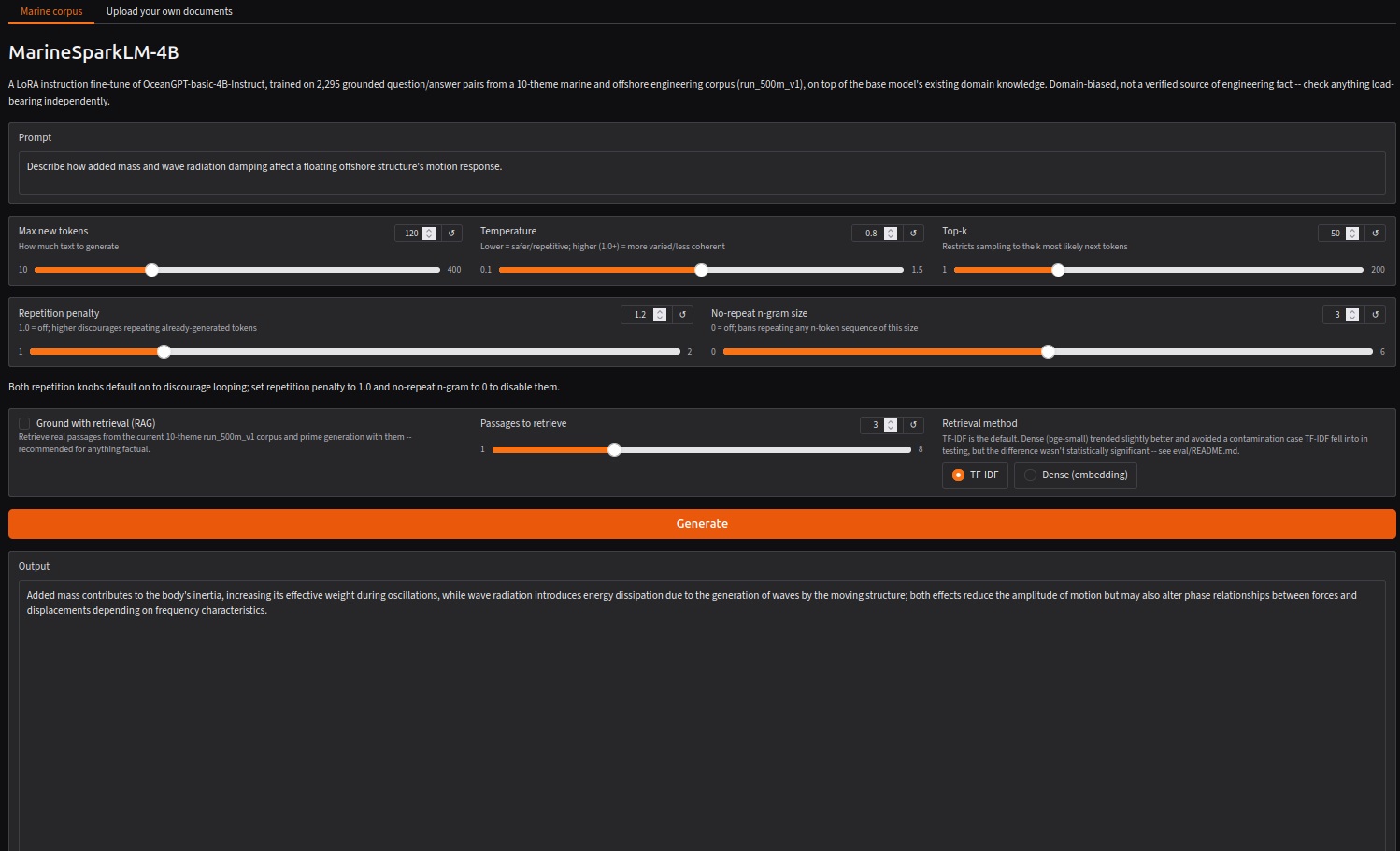}
\caption{Upper part of the web interface: prompt box, maximum new tokens (10 to 400, set to 120), temperature (0.1 to 1.5, set to 0.8), top-$k$ (1 to 200, set to 50), repetition penalty (1.0 to 2.0, set to 1.2), no-repeat n-gram size (0 to 6, set to 3), a retrieval switch and a passages slider (1 to 8, set to 3). The output panel, the retrieved-passages panel and the example prompts lie below the part shown.}
\end{figure}

A second tab lets users upload their own documents and grounds generation in those instead of the fixed corpus, following a retrieval-augmented-generation-with-small-backbone design: an uploaded file is chunked (1,200 characters, 150-character overlap) and embedded with bge-small-en-v1.5; a query retrieves the top 20 chunks by cosine similarity, a cross-encoder reranker (bge-reranker-base) reads query and passage together and keeps the top 3-5, and generation runs on the same base model and instruction adapter loaded in 4-bit NF4 rather than bf16 (3.1\,GB instead of 8.3\,GB, comparable quality, Section~\ref{sec:robustness}). The index is in-memory, per session, never persisted.

We validated this mode against the main corpus pipeline, with a caveat. 20 held-out questions were each run through it with that question's own real corpus passage as the simulated upload, and compared hand-scored against the existing main-corpus RAG generations for the identical 20 questions: upload RAG scored 91/120 (mean 4.55) against 72/120 (mean 3.60), mean difference $+0.95$, 95\% CI $[+0.25,+1.75]$, $p=0.0284$ (11 wins, 5 ties, 4 losses). This is not an apples-to-apples comparison: upload RAG only had to search the one document containing the answer, while the main pipeline searched the full 119,762-chunk corpus and could (and on one item did) retrieve an unrelated chunk instead, reproducing the keyword-collision contamination from Evaluation B. The result supports the design's premise but does not establish general superiority. Separately, a relevance floor on the reranker score, initially miscalibrated using reference-answer text, was recalibrated against real passages; 4 of 20 items still fell below it and generated ungrounded.

\end{document}